\documentclass[11pt]{article}
\usepackage{coling2020}
\usepackage{times}
\usepackage{url}
\usepackage{latexsym}

\usepackage{graphicx}
\usepackage{array}
\usepackage{amssymb}
\usepackage{amsfonts}
\usepackage{amsmath}
\usepackage{mathrsfs}
\usepackage{color}
\usepackage{booktabs}
\usepackage{threeparttable}
\usepackage{multirow}
\usepackage{times}
\usepackage{epsfig}
\usepackage{chngpage}
\usepackage{latexsym}
\usepackage{mathrsfs}
\usepackage{bm}
\usepackage{epstopdf}
\usepackage{tabularx}
\usepackage{hhline}
\usepackage{enumitem}
\usepackage{hyperref}
\usepackage{url}
\usepackage{booktabs}
\usepackage{subcaption}

\let\<\textless
\let\>\textgreater
\newcommand{\specialcell}[2][c]{%
  \begin{tabular}[#1]{@{}c@{}}#2\end{tabular}}

\newcommand{\la}{\left\langle}
\newcommand{\ra}{\right\rangle}
\newcommand{\lbar}{\left|}
\newcommand{\rbar}{\right|}
\newcommand{\ldot}{\left.}

\DeclareMathOperator*{\argmax}{arg\,max}

\colingfinalcopy % Uncomment this line for the final submission

\title{Quantum Inspired Word Representation and Computation}

\author{Shen Li \\
  {\tt \small{shen@mail.bnu.edu.cn}} \\\And
  Renfen Hu \\
  {\tt \small{irishu@mail.bnu.edu.cn}} \\\And
  Jinshan Wu \\
  {\tt \small{jinshanw@bnu.edu.cn}} \\\AND
  Beijing Normal University
}

\date{}

\begin{document}
\maketitle
\begin{abstract}
Word meaning has different aspects, while the existing word representation ``compresses'' these aspects into a single vector, and it needs further analysis to recover the information in different dimensions.
Inspired by quantum probability, we represent words as density matrices, which are inherently capable of representing mixed states. The experiment shows that the density matrix representation can effectively capture different aspects of word meaning while maintaining comparable reliability with the vector representation. Furthermore, we propose a novel method to combine the coherent summation and incoherent summation in the computation of both vectors and density matrices. It achieves consistent improvement on word analogy task.%\footnote{The source code in supplementary material will be released on GitHub.}
\end{abstract}

\section{Introduction}
\label{sec:introduction}

One of the most fundamental questions in Natural Language Processing is how to represent language units (words, sentences, texts) in a computational way, i.e. let the machine ``understand'' human language. The existing solutions evolve from symbolic expressions to mathematical vectors. 
Recently, the dense vector representation has achieved considerable success in various fields of NLP. The representative works include static word embeddings \cite{mikolov2013distributed,pennington2014glove} and dynamic contextual embeddings \cite{peters2018deep,devlin2018bert}.

When training word representations, the meaning of a word depends on its various contexts in a large corpus. Naturally, the word representation may have different aspects of properties by gaining information from different contexts. For example, the word ``king'' has high-level properties such as royalty and male gender \cite{levy2014linguistic}. This phenomenon could be easily observed from the nearest neighbors of its word embeddings.
However, the existing word representation ``compresses'' different aspects into a single vector, and it needs further analysis to recover the information in different dimensions \cite{rogers2018s}. The ``compression'' also exists in sentence representations. Obviously,  different extrinsic tasks may rely on different aspects of the representations. It naturally raises the following questions: Can we represent different aspects of language meaning inherently?

To address this issue, this paper introduces quantum probability into language representation, and attempts to represent words with density matrices. In physics, density matrix is used to describe a state of a particle. Each eigenstate of the particle is represented by a eigenvector in the density matrix, and each eigenvalue can be regarded as the probability of the particle in its corresponding eigenstate \cite{fano1957description}. Therefore, the density matrix representation is capable of storing mixed states. Meanwhile, it allows both coherent summation and incoherent summation of the eigenstates, which can be considered as vector summation and probability summation. 

After obtaining the word representation, we are interested in the computation between the representations. It is because the computation could illustrate the relations between words, and moreover, it is necessary for modeling higher level language units (sentences or texts) that are built upon the combination of word meaning in a specific way, e.g. Bag-of-words, CNNS, LSTMs and Transformers. If we can use vector summation and probability summation to combine different eigenstates of a word, can we apply them to the computation between words? Motivated by this question, we incorporate both vector summation and probability summation into computation of words.

%我们的方法可以用在各个模型上，这篇文章把word2vec选做了一个实验对象，是因为：1.word2vec广泛使用；2.word2vec是神经网络模型，在word2vec上的改进可以很容易推广到其他的模型。
%NLP的reviewer经常会质疑为什么不在最新的模型上改进，这点需要在写文章的时候就先回答。而且word2vec是2013年的模型了，和它对比意义不大，只是为了讲清楚density matrix才选用了它。
In the experiments, we implement our methods on skip-gram model with negative sampling (SGNS) \cite{mikolov2013distributed}, since it is one of the most popular models to train word representations. Meanwhile, as it is based on a simple yet effective neural network architecture, we could easily transfer the methods that work well in SGNS to other neural models. In addition, we impose the probability summation on both pre-trained vectors and pre-trained density matrices, as we will see from Equation (\ref{equation:analogy_vsps}). The effectiveness is testified by the analogy task, i.e. ``a is to b as c is to ?'' the solution of which traditionally relies on vector summation.

The experiments show the density matrix representation can effectively capture different aspects of word meaning while maintaining comparable reliability with the vector representation. Moreover, after further imposing probability summation, both the pre-trained vectors and density matrices achieve consistent improvement on word analogy tasks. It suggests that the probability summation could capture meaningful language information different from that of vector summation.
Thus the contribution of this paper is two-fold: 
\begin{itemize}
\item We introduce a novel word representation based on density matrix, and successfully apply it to neural network models. This representation can effectively capture different aspects of word meaning.
\item We integrate both probability summation and vector summation into the computation between words in both vector form and density matrix form. This computation method could better sketch the morphological and semantic relations between words.
\end{itemize}

\section{Theoretical background}
\label{sec:background}

Before introducing density matrix in quantum mechanics, we need to look at how to describe a state of a classical object.
In classical mechanics, objects can be in a mixture of states described by a probability distribution. 
For example, a coin can be in ``tail'' state (denoted as event $\lbar \psi_0 \ra = \lbar 0 \ra$) or ``head'' state (denoted as event $\lbar \psi_1 \ra = \lbar 1 \ra$), where $\lbar \psi \ra$, $\lbar 0 \ra$ and $\lbar 1 \ra$ are column vectors in Dirac notation \cite{dirac1939new}. Since the two states have nearly equal probability, the mixture state of a coin can be described as $\rho=\sum\limits_{i=0}^{1}p_{i}\lbar \psi_i \ra \la \psi_i \rbar=\frac{1}{2}\left(\lbar 0 \ra \la 0 \rbar + \lbar 1 \ra \la 1 \rbar\right)$, where $\la \psi \rbar$, $\la 0 \rbar$ and $\la 1 \rbar$ are row vectors. Here Dirac notation $\lbar 0 \ra \la 0 \rbar$ denotes the event that the coin is in state $0$. For classical object, neither $\lbar 0 \ra + \lbar 1 \ra$ nor $\lbar 0 \ra \la 1 \rbar$ is meaningful. That is to say, vector summation is meaningless for classical states, while probability summation is meaningful. In principle, there can be object's states derived from probability summations of simple events, e.g. the top of a coin $\rho=\sum\limits_{i=0}^{1}p_{i}\lbar \psi_i \ra \la \psi_i \rbar$ for any $p_{i}$ or the top of a dice $\rho=\sum\limits_{i=1}^{6}p_{i}\lbar \psi_i \ra \la \psi_i \rbar$ for any $p_{i}$. However, there is no such classical object whose states derived from $\lbar 0 \ra + \lbar 1 \ra$ or equivalently $\left(\lbar 0 \ra + \lbar 1 \ra\right)\left(\la 0 \rbar + \la 1 \rbar\right)= \lbar 0 \ra\la 0 \rbar + \lbar 1 \ra\la 1 \rbar+ \lbar 0 \ra\la 1 \rbar+ \lbar 1 \ra\la 0 \rbar$ (as we will see this equivalence later).

In quantum mechanics, quantum objects can be in a state derived from either vector summation or probability summation of simple events. In fact, the concept of simple events is slightly different from the classical simple events. All the pure states, which are the states involving only vector summation of simple events, such as $\lbar \psi_i \ra = \alpha_i \lbar 0 \ra + \beta_i \lbar 1 \ra$, form a vector space, so that any normalized orthogonal basis of the vector space can be seen as the set of simple events. This also means that a linear transformation of a set of chosen simple events (basis) is another set of simple events. The representation of classical objects is a particular situation in which only one of $\alpha$ and $\beta$ is not zero. Therefore, we can use the same mathematical form to represent mixture states of objects, i.e. density matrix
\begin{align}
  \rho = \sum_{ij}\rho_{ij}\lbar i \ra\la j\rbar \mbox{, or } \rho = \sum_i p_i \lbar \psi_i \ra \la \psi_i \rbar,
\end{align}    
as long as $\rho$ is hermitian, normalized and semi-positively defined, meaning correspondingly 
\begin{subequations}
\begin{align}
  \rho^{\dag} = \rho \Rightarrow \rho_{ij} = \rho^{*}_{ji}, \\
  tr\left(\rho\right)\triangleq \sum_{i} \la i\rbar \rho \lbar i \ra =1 \Rightarrow \sum_{i}\rho_{ii}=1, \\
  \rho_{\mu\mu}=\la \mu\rbar \rho \lbar \mu \ra\geq 0.
\end{align} 
\end{subequations}
Here in linear algebra terms, $\la \mu\rbar \ldot \nu \ra$ is the inner product of left (row) vector $\la \mu\rbar$ and the right (column) vector $\lbar \nu \ra$ (or of the right vectors $\lbar \mu\ra$ and $\lbar \nu \ra$), where if $\lbar \mu\ra$ and $\lbar \nu \ra$ belong to a normalized and orthogonal vector basis of the space then,
\begin{align}
  \la \mu\rbar \ldot \nu \ra =\delta_{\mu\nu},
\end{align} 
where $\delta_{\mu\nu}$ is the Kronecker delta;
$\la \mu\rbar \rho \lbar \mu \ra$ means a row vector $\la \mu\rbar$ times a matrix $\rho$ and then time a column vector $\lbar \mu \ra$, thus its end result is also a number; differently, the product of a column vector first and then a row vector, such as $\lbar i \ra\la j\rbar$, is in fact a matrix. Now we can see both the classical mixture states and quantum pure states can be represented in a density matrix form. 

It is experimental observation on quantum states that drives the theory of quantum systems to go beyond probability summation, and allow vector summation. In the current case of language phenomena, we assume that word meaning is a kind of mixed states, and both probability summation and vector summation of word representations are meaningful.

\section{Density matrix representation of words}
\label{sec:training}

\newtheorem{definition}{Definition}

Although it is not common to represent probabilities with negative values (eigenvalues in density matrix can be considered as probabilities), in fact, negative probabilities have been considered in quantum mechanics \cite{dirac1942bakerian,feynman1987negative}. Thus, in this paper, indefinite matrices are used to represent words instead of positive semi-definite matrices which are widely used in quantum probability. When negative eigenvalues are allowed in density matrix, it can give rise to more flexible representations. 

\subsection{Definitions}

Given space $\mathcal{L}^{n \times n}$, which belongs to Hilbert Space and consists of hermitian matrices, we have the following definitions on density matrix $\rho$. 

\begin{definition}
Given $\rho^A,\rho^B \in \mathcal{L}$, then the inner product is defined by 
\begin{align}
\la \rho^A, \rho^B \ra = Tr\left(\rho^{A{\dag}} \rho^B\right).
\end{align}
\label{definition:dot}
\end{definition}

\begin{definition}
Given $\rho^A \in \mathcal{L}$, then the norm is defined by 
\begin{align}
\lVert \rho^A \rVert = \sqrt{\la \rho^A, \rho^A \ra} = \sqrt{Tr\left(\left(\rho^A\right)^2\right)}.
\end{align}
\label{definition:length}
\end{definition}

\begin{definition}
Given $\rho^A,\rho^B \in \mathcal{L}$, then the distance between $\rho^A$ and $\rho^B$ is defined by 
\begin{align}
d_{\rho^A, \rho^B} = \lVert \rho^A - \rho^B \rVert.
\end{align}
\label{definition:euclid}
\end{definition}

\begin{definition}
Given $\rho^A,\rho^B \in \mathcal{L}$, then the cosine distance between $\rho^A$ and $\rho^B$ is defined by 
\begin{align}
cos\theta = \frac{\la \rho^A , \rho^B \ra}{\lVert \rho^A \rVert \lVert \rho^B \rVert} = \frac{Tr\left(\rho^{A{\dag}} \rho^B\right)}{\lVert \rho^A \rVert \lVert \rho^B \rVert}.
\end{align}
\label{definition:cosine}
\end{definition}

\subsection{Learning Word Representations}

In this paper, we implement our method in the skip-gram model with negative sampling (SGNS) \cite{mikolov2013distributed}, since it is one of the most popular models to train word representations. It should also be noted that SGNS is based on a simple yet effective neural network architecture. We could transfer the methods that work well in SGNS to other neural models.

To train word embeddings with SGNS, the optimization objective is 
\begin{align}
J = \argmax_{\theta} \sum_{\left(w,c\right) \in D}{\log{\sigma \left(v^c \cdot v^w\right)}} + \sum_{\left(w,c\right) \in D'}{\log{\sigma \left(-v^c \cdot v^w \right) }},
\end{align}
where $\theta$ denotes the parameters of all vectors $v$, $w$ and $c$ are central word and context word respectively, $v^w$ and $v^c$ are their embeddings, $\sigma$ is sigmoid function and $(w,c) \in D$ means that $w$ and $c$ cooccur in a window in the document $D$ and $(w,c) \in D'$ represents that two words cooccur in Negative Sampling.

The objective can be rewritten as 
\begin{align}
J = \argmax_{\theta} \sum_{\left( w,c \right)  \in D}{\log{\sigma \left( \la v^c , v^w \ra \right) }} + \sum_{\left( w,c \right)  \in D'}{\log{\sigma \left( - \la v^c , v^w \ra \right) }}.
\end{align}

If words are represented by density matrices, then the objective becomes
\begin{align}
\begin{split}
J 
&= \argmax_{\theta} \sum_{\left( w,c \right)  \in D}{\log{\sigma \left( \la \rho^c , \rho^w \ra \right) }} + \sum_{\left( w,c \right)  \in D'}{\log{\sigma \left( - \la \rho^c , \rho^w \ra \right) }} \\
&= \argmax_{\theta} \sum_{\left( w,c \right)  \in D}{\log{\sigma \left( Tr\left( \rho^{w{\dag}} \rho^c \right)  \right) }} + \sum_{\left( w,c \right)  \in D'}{\log{\sigma \left( - Tr\left( \rho^{w{\dag}} \rho^c \right)  \right) }}.
\end{split}
\end{align}

In word2vec, to update parameters in SGNS, the process is
\begin{align}
\theta\left( t+1 \right)  = \theta\left( t \right)  - \eta \frac{\partial J}{\partial \theta\left( t \right) },
\end{align}
where $t$ is the number of steps and $\eta$ is the learning rate.
Taking a central word $w$ and a context word $c \in D \cup D'$ as an example, to update $v^w$, the process can be written like
\begin{align}
\begin{split}
v^w\left( t+1 \right)  
&= v^w\left( t \right)  - \eta \frac{\partial J}{\partial v^w\left( t \right) } \\
&= v^w\left( t \right)  + \eta \left( l - \sigma\left( \la v^w\left( t \right) , v^c\left( t \right)  \ra \right)  \right)  v^c\left( t \right)  \\
&= v^w\left( t \right)  + \eta \left( l - \sigma\left(  v^w\left( t \right) v^c\left( t \right)   \right)  \right)  v^c\left( t \right)  \\
&= v^w\left( t \right)  + \eta \left( l - \sigma\left(  \sum_{i}{v^w_i\left( t \right) v^c_i\left( t \right) }  \right)  \right)  v^c\left( t \right),
\end{split}
\end{align}
where $l$ is $1$ if $c \in D$ or $0$ if $c \in D'$.

If words are represented by hermitian matrices, the equation becomes
\begin{align}
\begin{split}
\rho^w\left( t+1 \right)  
&= \rho^w\left( t \right)  + \eta \left( l - \sigma\left( \la \rho^w\left( t \right) , \rho^c\left( t \right)  \ra \right)  \right)  \rho^c\left( t \right)  \\
&= \rho^w\left( t \right)  + \eta \left( l - \sigma\left( Tr\left( \rho^w\left( t \right) ^{\dag}\rho^c\left( t \right)  \right)  \right)  \right)  \rho^c\left( t \right)  \\
&= \rho^w\left( t \right)  + \eta \left( l - \sigma\left( Tr\left( \rho^w\left( t \right) \rho^c\left( t \right)  \right)  \right)  \right)  \rho^c\left( t \right)  \\
&= \rho^w\left( t \right)  + \eta \left( l - \sigma\left( \sum_{ij}{\rho^w_{ij}\left( t \right) \rho^c_{ji}\left( t \right) } \right)  \right)  \rho^c\left( t \right).
\end{split}
\end{align}

For instance, given
\begin{subequations}
\begin{align}
v^w\left( t \right) =\left(\begin{matrix}a^w_0\ a^w_1\ a^w_2\ b^w_1\end{matrix}\right)\mbox{, }
v^c\left( t \right) =\left(\begin{matrix}a^c_0\ a^c_1\ a^c_2\ b^c_1\end{matrix}\right),
\end{align}
\begin{align}
\rho^w\left( t \right) =
\left(
  \begin{matrix}
    {a^w_0}\ \ \ \ & {a^w_1 + b^w_1i} \\
    {a^w_1 - b^w_1i}\ & {a^w_2}\ \ \   \\
  \end{matrix} 
\right)\mbox{, }
\rho^c\left( t \right) =
\left(
  \begin{matrix}
    {a^c_0}\ \ \ \ & {a^c_1 + b^c_1i} \\
    {a^c_1 - b^c_1i}\ & {a^c_2}\ \ \   \\
  \end{matrix} 
\right),
\end{align}
\end{subequations}
then
\begin{subequations}
\begin{align}
\begin{split}
\la v^w\left( t \right) , v^c\left( t \right)  \ra
&=\left(\begin{matrix}a^w_0\ a^w_1\ a^w_2\ b^w_1\end{matrix}\right) \left(\begin{matrix}a^c_0\ a^c_1\ a^c_2\ b^c_1\end{matrix}\right) \\
&=a^w_0a^c_0 + a^w_1a^c_1 + b^w_1b^c_1 + a^w_2a^c_2,
\end{split}
\end{align}
\begin{align}
\begin{split}
\la \rho^w\left( t \right)  , \rho^c\left( t \right)  \ra 
&=Tr\left( \rho^w\left( t \right) ^{\dag}\rho^c\left( t \right)  \right)  \\
&=
\left(
  \begin{matrix}
    {a^w_0}\ \ \ \ & {a^w_1 + b^w_1i} \\
    {a^w_1 - b^w_1i}\ & {a^w_2}\ \ \   \\
  \end{matrix} 
\right) 
\left(
  \begin{matrix}
    {a^c_0}\ \ \ \ & {a^c_1 + b^c_1i} \\
    {a^c_1 - b^c_1i}\ & {a^c_2}\ \ \   \\
  \end{matrix} 
\right) \\
&=a^w_0a^c_0 + 2\left( a^w_1a^c_1 + b^w_1b^c_1 \right)  + a^w_2a^c_2.
\end{split}
\end{align}
\label{equation:compare}
\end{subequations}
It can be seen that two kinds of inner product are similar. The inner product of density matrices has more items coming from off-diagonal elements.
Thus, it is easy to extend SGNS to support density matrix by double off-diagonal elements in computation of inner product.

\subsection{Performance of the density matrix representation }
\label{sec:Performance}

\begin{table}[ht!]
\begin{center}
\begin{tabular}{|c|c|c|c|c|}
\hline
Window & Iteration & Sub-sampling & \specialcell{Low-frequency \\threshold} & Negative Sampling \\
\hline
5      & 5         & 1e-3         & 20                      & 10       \\
\hline
\end{tabular}
\end{center}
\caption{Hyperparameters in the training of word matrices.}
\label{table:hyperparameters}
\end{table}

\begin{table}[ht!]
\begin{small}
\begin{center}
\setlength\tabcolsep{4px}
\begin{tabular}{|c||c|c||c|c||c|c||c|c|}
\hline
Dataset & \specialcell{Vector \\ $dim=36$} & \specialcell{Matrix \\ $dim=8$} & \specialcell{Vector \\ $dim=136$} & \specialcell{Matrix \\ $dim=16$} & \specialcell{Vector \\ $dim=300$} & \specialcell{Matrix \\ $dim=24$} & \specialcell{Vector \\ $dim=528$} & \specialcell{Matrix \\ $dim=32$} \\
\hline
\specialcell{Google\\syntactic} & 32.85\% & 33.33\% & 62.03\% & 61.23\% & 64.59\% & 65.55\% & 64.68\% & 65.36\% \\
\hline
\specialcell{Google\\semantic} & 34.18\% & 36.64\% & 66.13\% & 66.79\% & 74.79\% & 74.68\% & 76.05\% & 74.82\% \\
\hline
\specialcell{BATS\\Inflectional\\morphology} & 35.32\% & 36.27\% & 59.25\% & 59.49\% & 61.11\% & 62.28\% & 60.30\% & 60.80\% \\
\hline
\specialcell{BATS\\Derivational\\morphology} & 5.28\% & 5.16\% & 11.47\% & 12.02\% & 12.74\% & 12.81\% & 11.47\% & 11.98\% \\
\hline
\specialcell{BATS\\Encyclopedic\\semantics} & 16.74\% & 17.79\% & 34.71\% & 34.74\% & 38.50\% & 38.79\% & 37.36\% & 38.73\% \\
\hline
\specialcell{BATS\\Lexicographic\\semantics} & 4.11\% & 5.28\% & 9.09\% & 10.56\% & 9.97\% & 10.56\% & 9.38\% & 10.56\% \\
\hline
\end{tabular}
\end{center}
\end{small}
\caption{The results of word analogy task with different dimensions. For fair comparison, the number of parameters of vector form is the same as that of density matrix in each group. For instance, the number of parameters of 300 dimensional vector is equal to that of 24 dimensional density matrix.}
\label{table:analogy}
\end{table}

\begin{table}
\begin{subtable}[t]{1.0\linewidth}
\begin{small}
\begin{center}
\setlength\tabcolsep{3px}
\begin{tabular}{|c|c|c|c|c|c|c|c|c|}
\hline
\multirow{2}{*}{$\rho$} & \multicolumn{8}{c|}{Eigenvalues} \\
  \hhline{~--------}
  & -0.6330 & -0.3572 & -0.2969 & -0.0917 & 0.0119 & 0.1852 & 0.3605 & 0.4591 \\
\hline
banks & receivership & equatorial & denmark's & campbells & anzac & 0.08 & integrity & gazprom \\ 
\hline
bank's & liquidation & china & sweden's & taylors & 42,000 & 1.13 & plaintiff's & bank's \\ 
\hline
citibank & bankrupt & yucatn & biggest & wayside & 111th & 0.06 & deeds & markov \\ 
\hline
barclays & scrapping & cocos & seti & friends & 29,000 & 0.30 & upholding & putin \\ 
\hline
hsbc & parcels & yucatan & kingdom's & clarks & 95th & 1.16 & claim & nasdaq \\ 
\hline
trading & thoroughbreds & formosa & kuru & grandfathers & regiment's & 0.16 & denying & ticker \\ 
\hline
savings & distilleries & verde & aral & boyhood & 99th & 0.12 & overturning & telescope \\ 
\hline
lloyds & nationalised & suzhou & norway's & mungo & 52nd & 0.07 & petitioners & fsb \\ 
\hline
capital & leases & mexico & dg & ebenezer & 79th & 0.13 & affirmed & subscriber \\ 
\hline
banking & franchised & peru & europe's & david's & rhodesian & 0.09 & asserting & operator \\ 
\hline
depository & sale & sao & europes & ridgeway & 33,000 & 0.10 & manorial & post's \\ 
\hline
deposit & freight & central & cementing & benefactors & wales's & 0.54 & aforesaid & hubble \\ 
\hline
banque & government-owned & sumatra & britain's & nether & 103rd & 0.20 & usury & telegram \\ 
\hline
citigroup & tramways & kunming & aga & kinsmen & 48,000 & 0.38 & assessor & ftse \\ 
\hline
river & plying & phong & sokka & doon & eritrean & 0.11 & dissented & wachovia \\ 
\hline
parcel & furlongs & biscayne & rana & gwyn & 7,500 & 0.21 & lott & observer \\ 
\hline
abn & bankruptcy & xiamen & asia's & edmund's & gurkha & 0.28 & infallible & sputnik \\ 
\hline
canal & subsidiaries & seaport & natwest & mavis & 21,000 & 1.23 & tithes & kgb \\ 
\hline
brokerage & dealerships & dali & saga & parson & 98th & 0.43 & imposing & nis \\
\hline
\end{tabular}
\end{center}
\end{small}
\caption{Nearest neighbors of \emph{bank}.}
\label{table:neighbors_bank}
\end{subtable}

\begin{subtable}{1.0\linewidth}
\begin{small}
\begin{center}
\setlength\tabcolsep{2px}
\begin{tabular}{|c|c|c|c|c|c|c|c|c|}
\hline
\multirow{2}{*}{$\rho$} & \multicolumn{8}{c|}{Eigenvalues} \\
  \hhline{~--------}
   &  -0.5497 & -0.4772 & -0.2473 & -0.1119 & -0.0157 & 0.1921 & 0.3008 & 0.5184 \\
\hline
science-fiction & explored & comedies & whirlpool & carving & 1603 & ettore & rotten & astronaut \\
\hline
non-fiction & exploring & novelists & eau & vase & 1604 & mle & inaccuracies & observer \\
\hline
nonfiction & spin-offs & ziegfeld & racetrack & dinghy & 1651 & ddr & tracts & skeptic \\
\hline
novels & adv & beauties & lode & archery & 1599 & luigi & negatives & watcher \\
\hline
anthology & evolutions & satirized & aqueduct & blades & 1584 & renzo & plagiarism & creationist \\
\hline
thrillers & researching & hollywood's & mists & springboard & 1578 & e5 & libel & airman \\
\hline
comic & viz & follies & euclid & finish & 1597 & brno & seditious & shooter \\
\hline
mystery & documenting & operettas & woodbine & teak & 1707 & steyr & forgeries & seeker \\
\hline
novel & independently & burlesque & brea & silverware & 1593 & bugatti & clippings & naismith \\
\hline
bestselling & revisited & churchill's & raceway & round & 1671 & 35mm & allegations & sailor \\
\hline
anthologies & spawned & revues & biel & leg & 1644 & krzysztof & op-ed & horseman \\
\hline
asimov's & recent & wholesome & haute & quadrangular & 1661 & dv & obscenity & communicator \\
\hline
comics & adapting & actresses & chasm & berth & 1642 & glock & unsubstantiated & creationism \\
\hline
heinlein & extensively & parodying & hairpin & bowls & 1662 & d5 & antiquities & atheist \\
\hline
cyberpunk & predating & heroines & downhill & putt & 1669 & dac & fingerprints & athlete \\
\hline
dystopian & episodic & musicals & descartes & spectacularly & 1592 & carlo & archival & mountaineer \\
\hline
suspense & aspects & astaire & valhalla & lancers & 1606 & bosch & evidence & starfleet \\
\hline
stories & adaptations & baum's & blvd & turf & 1724 & riccardo & misrepresentation & hawkeye \\
\hline
fandom & interrelated & wilde's & edina & jetty & 1601 & c5 & accusations & sportswriter \\
\hline
comic-book & latest & entertainments & seca & canoes & dafydd & bolzano & sedition & parallax \\
\hline
\end{tabular}
\end{center}
\end{small}
\caption{Nearest neighbors of \emph{fiction}.}
\label{table:neighbors_fiction}
\end{subtable}
\caption{Nearest neighbors of \emph{bank} and \emph{fiction} according to density matrix $\rho$ and different eigenvectors of the matrices.}
\label{table:neighbors}
\end{table}

In the experiments, we train word vectors and word matrices \footnote{Because of Equation (\ref{equation:compare}), the inner product could not be affected by the introduction of imaginary part. Thus, we use density matrices with real numbers in this paper for a fair comparison with word2vec.} with the same number of parameters on Wiki2010 corpus \footnote{Download from \url{http://nlp.stanford.edu/data/WestburyLab.wikicorp.201004.txt.bz2} and clean it with the same strategy in \cite{levy2015improving}}. The settings can be seen in Table \ref{table:hyperparameters}. To testify the reliability of word representations, we evaluate the vectors and matrices on word analogy datasets BATS \cite{gladkova2016analogy} and Google Analogy \cite{mikolov2013efficient}.

As shown in Table \ref{table:analogy}, density matrices and vectors (with the same number of parameters) achieve similar accuracies consistently from low to high dimensions. It indicates that matrices can represent what vectors encode. Equation (\ref{equation:compare}) can easily explain the experiment results. If off-diagonal elements of density matrices time $\frac{1}{\sqrt{2}}$, these two methods have similar effects in computation of inner product.

As we all know, given a density matrix $\rho$, it can be written as the combination of its eigenvectors, $\rho = \sum i \lbar i \ra \la i \rbar$, where $i$ is an eigenvalue and $\lbar i \ra$ is its corresponding eigenstate. As eigenvectors of a density matrix can represent pure states, we could examine the nearest neighbors of each pure state by using $\lbar i \ra \la i \rbar$ if its eigenvalue $i \geqslant 0$ or $-\lbar i \ra \la i \rbar$ if its eigenvalue $i < 0$ rather than $\rho$. 
$-\lbar i \ra \la i \rbar$ is the natural result when we allow negative probabilities in density matrices. As a reference, we also retrieve the nearest neighbors of $\rho$. The similarity is based on cosine distance (Definition \ref{definition:cosine}).

As shown in Table \ref{table:neighbors}, we find that the eigenvectors in density matrices could effectively capture different aspects of word meaning, and the absolute eigenvalues are also meaningful. Eigenvectors with larger values have greater impact on word meaning than those with smaller values. Table \ref{table:neighbors_bank} shows an example of the word \emph{bank}. The first column is the nearest neighbors of $\rho$ which represents the whole word meaning, and the other columns are the nearest neighbors of each eigenvector.
If we use $\rho$ to find the nearest neighbors, the neighbors of ``bank'' in different senses are mixed together. But if we use $\lbar i \ra \la i \rbar$ (or $-\lbar i \ra \la i \rbar$), the nearest neighbors can well reflect different aspects of word meaning. For instance, the words in the first three columns and the last two columns have relatively high absolute eigenvalues. The words in first three columns are related with commercial bank, location and bank names respectively, while the words in the last two columns are mostly judicial and financial terms. The similar phenomenon can also be found in Table \ref{table:neighbors_fiction} regarding the word \emph{fiction}. The first column is related with fiction content, the second column is about drama and movie, and the last column involves characters.

\section{Computation with probability summation and vector summation}
\label{sec:psvs}

The computation method between words plays crucial roles in natural language representation, since it could sketch the relations between words, and serve as a basis for modeling higher level language units such as sentences or texts. As we introduced in Section \ref{sec:background}, probability summation and vector summation are two important forms of computation in quantum probability. However, the existing vector representation utilizes only vector summation. In this study, we impose the probability summation on both pre-trained vectors and pre-trained density matrices. Here we illustrate this method with analogy task as an example. 

In an analogy question e.g. $king - man + woman \approx queen$, we have 4 word vectors $\lbar a \ra$, $\lbar b \ra$, $\lbar c \ra$ and $\lbar d \ra$ accordingly. The process can be written as
\begin{align}
\lbar x \ra = \lbar b \ra - \lbar a \ra + \lbar c \ra.
\label{equation:analogy_vector}
\end{align}
If the nearest neighbor of $\lbar x \ra$ is $\lbar d \ra$, $\lbar x \ra$ is the correct answer. If not, $\lbar x \ra$ is a wrong answer.

We can transform vector summation Equation (\ref{equation:analogy_vector}) into
\begin{align}
\lbar x \ra \la x \rbar = (\lbar b \ra - \lbar a \ra + \lbar c \ra)(\la b \rbar - \la a \rbar + \la c \rbar).
\label{equation:analogy_vs}
\end{align}
According to Section \ref{sec:background}, the probability summation can be written as
\begin{align}
\lbar x \ra \la x \rbar = \lbar b \ra \la b \rbar - \lbar a \ra \la a \rbar + \lbar c \ra \la c \rbar.
\label{equation:analogy_ps}
\end{align}
After combining Equation (\ref{equation:analogy_vs}) and Equation (\ref{equation:analogy_ps}), we can get 
\begin{align}
\begin{split}
\lbar x \ra \la x \rbar = (\lbar b \ra - \lbar a \ra + \lbar c \ra)(\la b \rbar - \la a \rbar + \la c \rbar) + \alpha (\lbar b \ra \la b \rbar - \lbar a \ra \la a \rbar + \lbar c \ra \la c \rbar),
\end{split}
\label{equation:analogy_vsps}
\end{align}
where $\alpha$ is a factor to balance the ratio of vector summation to probability summation. A larger $\alpha$ denotes higher weight of probability summation in the computation. 

We can also easily apply this method to density matrix representations of words, 
\begin{align}
\begin{split}
\rho^x \otimes \rho^x = (\rho^b - \rho^a + \rho^c) \otimes (\rho^b - \rho^a + \rho^c) + \alpha (\rho^b \otimes \rho^b - \rho^a \otimes \rho^a + \rho^c \otimes \rho^c),
\end{split}
\label{equation:analogy_vsps_matrix}
\end{align}
where $\otimes$ means Kronecker product.

\begin{table}
\begin{small}
\begin{center}
\begin{tabular}{|c|c|c|c|c|c|c|c|}
\hline
Type & $\alpha$ & \specialcell{Google\\syntactics} & \specialcell{Google\\semantics} & \specialcell{BATS\\Inflectional\\morphology} & \specialcell{BATS\\Derivational\\morphology} & \specialcell{BATS\\Encyclopedic\\semantics} & \specialcell{BATS\\Lexicographic\\semantics}\\
\hline
\multirow{13}{*}{\specialcell{Vector \\ $dim=36$}} 
& \underline{0.0} & \underline{32.85\%} & \underline{34.18\%} & \underline{35.32\%} & \underline{5.28\%} & \underline{16.74\%} & \underline{4.11\%} \\
\hhline{~-------}
& 0.1 & 33.52\% & 34.46\% & 36.07\% & 5.40\% & 16.78\% & 4.69\% \\
\hhline{~-------}
& 0.2 & 33.77\% & 34.71\% & 36.51\% & 5.62\% & 16.67\% & 4.99\% \\
\hhline{~-------}
& 0.3 & 34.15\% & 34.85\% & 36.65\% & 5.70\% & 16.54\% & 5.28\% \\
\hhline{~-------}
& 0.4 & 34.46\% & 35.03\% & \textbf{36.91\%} & 5.81\% & 16.72\% & 5.28\% \\
\hhline{~-------}
& 0.5 & 34.49\% & 34.98\% & 36.83\% & 5.87\% & 16.87\% & 5.28\% \\
\hhline{~-------}
& 0.6 & \textbf{34.58\%} & 35.09\% & 36.78\% & 5.95\% & 16.99\% & 5.28\% \\
\hhline{~-------}
& 0.7 & 34.53\% & 35.05\% & 36.75\% & 5.88\% & 16.97\% & 5.28\% \\
\hhline{~-------}
& 0.8 & 34.53\% & \textbf{35.12\%} & 36.76\% & 5.90\% & 17.01\% & 5.57\% \\
\hhline{~-------}
& 0.9 & 34.41\% & 35.03\% & 36.66\% & \textbf{5.97\%} & \textbf{17.03\%} & 5.87\% \\
\hhline{~-------}
& 1.0 & 34.26\% & 34.94\% & 36.50\% & \textbf{5.97\%} & 16.99\% & \textbf{6.16\%} \\
\hhline{~-------}
& 1.1 & 34.30\% & 34.98\% & 36.35\% & 5.94\% & 16.99\% & \textbf{6.16\%} \\
\hhline{~-------}
& 1.2 & 34.20\% & 34.98\% & 36.29\% & \textbf{5.97\%} & 16.99\% & \textbf{6.16\%} \\
\hhline{|=|=|=|=|=|=|=|=|}
\multirow{13}{*}{\specialcell{Vector \\ $dim=300$}} 
& -0.5 & 66.19\% & 73.32\% & 64.31\% & 13.28\% & 37.59\% & 8.80\% \\
\hhline{~-------}
& -0.4 & 66.71\% & 74.30\% & \textbf{64.85\%} & 14.00\% & 38.80\% & 9.09\% \\
\hhline{~-------}
& -0.3 & \textbf{66.92\%} & 74.73\% & 64.12\% & \textbf{14.01\%} & \textbf{38.95\%} & 9.09\% \\
\hhline{~-------}
& -0.2 & 66.46\% & 74.71\% & 63.23\% & 13.65\% & 38.84\% & 9.68\% \\
\hhline{~-------}
& -0.1 & 65.75\% & \textbf{74.80\%} & 62.17\% & 13.26\% & 38.77\% & 9.97\% \\
\hhline{~-------}
& \underline{0.0} & \underline{64.59\%} & \underline{74.79\%} & \underline{61.11\%} & \underline{12.74\%} & \underline{38.50\%} & \underline{9.97\%} \\
\hhline{~-------}
& 0.1 & 63.80\% & 74.61\% & 60.10\% & 12.07\% & 38.19\% & \textbf{10.56\%} \\
\hhline{~-------}
& 0.2 & 63.05\% & 74.35\% & 59.30\% & 11.56\% & 37.78\% & \textbf{10.56\%} \\
\hhline{~-------}
& 0.3 & 62.46\% & 74.30\% & 58.44\% & 11.13\% & 37.43\% & \textbf{10.56\%} \\
\hhline{~-------}
& 0.4 & 61.60\% & 73.89\% & 57.55\% & 10.71\% & 37.05\% & \textbf{10.56\%} \\
\hhline{~-------}
& 0.5 & 60.96\% & 73.58\% & 56.75\% & 10.47\% & 36.64\% & \textbf{10.56\%} \\
\hhline{~-------}
& 0.6 & 60.44\% & 73.27\% & 56.06\% & 10.22\% & 36.26\% & \textbf{10.56\%} \\
\hhline{~-------}
& 0.7 & 59.78\% & 72.89\% & 55.56\% & 9.92\% & 35.88\% & \textbf{10.56\%} \\
\hhline{|=|=|=|=|=|=|=|=|}
\multirow{13}{*}{\specialcell{Density \\ matrix \\ $dim=8$}} 
& \underline{0.0} & \underline{33.33\%} & \underline{36.64\%} & \underline{36.27\%} & \underline{5.16\%} & \underline{17.79\%} & \underline{5.28\%} \\
\hhline{~-------}
& 0.1 & 34.04\% & 36.92\% & 36.62\% & 5.22\% & 18.18\% & 5.57\% \\
\hhline{~-------}
& 0.2 & 34.40\% & 36.99\% & 36.90\% & 5.34\% & 18.38\% & 5.87\% \\
\hhline{~-------}
& 0.3 & 34.63\% & 37.20\% & \textbf{36.98\%} & 5.47\% & 18.40\% & 6.16\% \\
\hhline{~-------}
& 0.4 & 34.95\% & 37.15\% & 36.91\% & 5.57\% & 18.40\% & 6.16\% \\
\hhline{~-------}
& 0.5 & 35.04\% & 37.26\% & 36.86\% & 5.61\% & 18.35\% & 6.16\% \\
\hhline{~-------}
& 0.6 & \textbf{35.23\%} & 37.22\% & 36.86\% & 5.66\% & 18.35\% & 6.45\% \\
\hhline{~-------}
& 0.7 & 35.21\% & 37.28\% & 36.73\% & 5.67\% & 18.42\% & 6.74\% \\
\hhline{~-------}
& 0.8 & 35.18\% & 37.31\% & 36.61\% & 5.72\% & 18.40\% & 6.74\% \\
\hhline{~-------}
& 0.9 & 35.15\% & \textbf{37.35\%} & 36.56\% & 5.80\% & \textbf{18.44\%} & \textbf{7.33\%} \\
\hhline{~-------}
& 1.0 & 35.12\% & 37.24\% & 36.39\% & \textbf{5.91\%} & 18.38\% & \textbf{7.33\%} \\
\hhline{~-------}
& 1.1 & 35.10\% & 37.14\% & 36.21\% & 5.86\% & 18.31\% & \textbf{7.33\%} \\
\hhline{~-------}
& 1.2 & 35.03\% & 37.11\% & 36.05\% & 5.86\% & 18.33\% & \textbf{7.33\%} \\
\hhline{|=|=|=|=|=|=|=|=|}
\multirow{13}{*}{\specialcell{Density \\ matrix \\ $dim=24$}} 
& -0.5 & 66.67\% & 73.86\% & 65.45\% & 13.68\% & 38.24\% & 9.38\% \\
\hhline{~-------}
& -0.4 & \textbf{67.61\%} & 74.53\% & \textbf{65.62\%} & \textbf{14.42\%} & 38.95\% & 9.38\% \\
\hhline{~-------}
& -0.3 & 67.30\% & \textbf{74.71\%} & 65.10\% & 14.17\% & \textbf{39.29\%} & 9.38\% \\
\hhline{~-------}
& -0.2 & 66.91\% & 74.70\% & 64.37\% & 13.69\% & 39.18\% & 9.68\% \\
\hhline{~-------}
& -0.1 & 66.33\% & 74.70\% & 63.19\% & 13.36\% & 39.04\% & 10.26\% \\
\hhline{~-------}
& \underline{0.0} & \underline{65.55\%} & \underline{74.68\%} & \underline{62.28\%} & \underline{12.81\%} & \underline{38.79\%} & \underline{\textbf{10.56\%}} \\
\hhline{~-------}
& 0.1 & 64.65\% & 74.30\% & 61.19\% & 12.20\% & 38.70\% & \textbf{10.56\%} \\
\hhline{~-------}
& 0.2 & 63.76\% & 74.16\% & 60.28\% & 11.63\% & 38.30\% & \textbf{10.56\%} \\
\hhline{~-------}
& 0.3 & 63.10\% & 73.92\% & 59.37\% & 11.14\% & 37.88\% & \textbf{10.56\%} \\
\hhline{~-------}
& 0.4 & 62.26\% & 73.69\% & 58.42\% & 10.85\% & 37.50\% & 10.26\% \\
\hhline{~-------}
& 0.5 & 61.59\% & 73.47\% & 57.66\% & 10.52\% & 36.94\% & 10.26\% \\
\hhline{~-------}
& 0.6 & 60.89\% & 73.14\% & 57.02\% & 10.16\% & 36.76\% & 10.26\% \\
\hhline{~-------}
& 0.7 & 60.29\% & 72.92\% & 56.39\% & 9.90\% & 36.17\% & 10.26\% \\
\hline
\end{tabular}
\end{center}
\end{small}
\vskip -0.2cm
\caption{The results of word analogy task with different $\alpha$. Larger $\alpha$ indicates larger weight of probability summation. The pre-trained vectors and density matrices come from Section \ref{sec:Performance}.}
\label{table:analogy_a}
\end{table}

\begin{table}
\begin{subtable}[t]{1.0\linewidth}
\begin{small}
\begin{center}
\setlength\tabcolsep{4px}
\begin{tabular}{|c|c|c|c|c|c|}
\hline
\multicolumn{2}{|c|}{$deaths - death + street$} & \multicolumn{2}{c|}{$laws - law + student$} & \multicolumn{2}{c|}{$colonization - colonize + condense$} \\
\hline
$\alpha = -0.3$ & $\alpha = 0$ & $\alpha = -0.3$ & $\alpha = 0$ & $\alpha = -0.3$ & $\alpha = 0$\\
\hline
streets & campgrounds & students & ylc & condensation & solidification\\
\hline
avenues & carparks & ylc & cross-campus & condenses & vacuo\\
\hline
bergenline & avenues & non-student & nwsa & crystallization & crystallization\\
\hline
mid-block & ne/sw & cross-campus & nacw & solidification & adiabatic\\
\hline
woodhaven & e/w & non-fraternity & cluw & vacuo & photochemical\\
\hline
cambie & n/s & ex-students & students & photochemical & condensation\\
\hline
n/s & streets & nwsa & ncjw & gaseous & distillation\\
\hline
akard & cross-streets & nacw & alsf & adiabatic & calcination\\
\hline
e/w & lakefront & icfj & non-fraternity & condensing & sublimation\\
\hline
lakefront & trailheads & agbu & icfj & calcination & condenses\\
\hline
\end{tabular}
\end{center}
\end{small}
\subcaption{Comparison in vector form ($dim=300$).}
\label{table:analogy_case_vector}
\end{subtable}

\begin{subtable}[t]{1.0\linewidth}
\begin{small}
\begin{center}
\setlength\tabcolsep{4px}
\begin{tabular}{|c|c|c|c|c|c|}
\hline
\multicolumn{2}{|c|}{$on - off + west$} & \multicolumn{2}{c|}{$harassment - harass + invest$} & \multicolumn{2}{c|}{$configuration - configure + derive$} \\
\hline
$\alpha = -0.3$ & $\alpha = 0$ & $\alpha = -0.3$ & $\alpha = 0$ & $\alpha = -0.3$ & $\alpha = 0$\\
\hline
east & kingshighway & investment & equity & derivation & armenoceras\\
kingshighway & boxborough & equity & outlays & deriving & disconformity\\
north & 2427 & financing & mdri & frw & imbrian\\
gainsboro & holmfield & investing & employer-sponsored & folium & folium\\
south & fredericton's & outlays & medicare & armenoceras & frw\\
northcliff & burnaby's & policyholders & cwsrf & derived & derivation\\
burnaby's & gainsboro & employer-sponsored & policyholders & concordant & actinoceras\\
kenthurst & section & profit & unfunded & backstaff & natrolite\\
holmfield & 26-28 & private-sector & calpers & achromat & end-member\\
clareville & kenthurst & benefits & hythiam & disconformity & mordent\\
\hline
\end{tabular}
\end{center}
\end{small}
\subcaption{Comparison in density matrix form ($dim=24$).}
\label{table:analogy_case_matrix}
\end{subtable}
\caption{Different answers and candidates of analogical questions when factors $\alpha$ are set different values. Words are arranged from top to bottom according to their scores. The top one has the largest score.}
\label{table:analogy_case}
\end{table}

Table \ref{table:analogy_a} shows the results with different weight factor $\alpha$. When $\alpha$ is 0, the computation allows only vector summation, thus it is as same as those in Table \ref{table:analogy}. 
It can be seen that after introducing probability summation, the performances are consistently improved in both vector and density matrix forms. It indicates that both summation could leverage unique information that is ignored by the other. 
Since only the computation method is changed here, and all the vectors and matrices are pre-trained, the results could not be affected by external factors (e.g. random seed) but only the combination of probability summation and vector summation.
Therefore, probability summation could help the representations better capture the morphological and semantic relations between words.

To better understand the results, Table \ref{table:analogy_case} lists several cases comparing pure vector summation ($\alpha = 0$) and mixed summation ($\alpha = -0.3$). We can easily find the positive changes introduced by probability summation. For instance, in Table \ref{table:analogy_case_vector}, the answer of the second question is ``students'', but there are some mixture of words related to students and university abbreviations when only vector summation is used. After adding probability summation, words related to students are closer to the correct answer of the question and they are nearly not mixed with university abbreviations. The similar phenomenon happens in the first column of Table \ref{table:analogy_case_matrix}, where the locality nouns ``east'', ``north'' and ``south'' all get closer to the target.

\section{Related work}
\label{sec:relatedwork}

There are several works which leverage density matrix to enhance language representations.
\newcite{sordoni2013modeling} transform one-hot representation of a sentence into density matrix (quantum language model, QLM) with $R \rho R$ algorithm, and it outperforms classical models on ad-hoc retrieval tasks. 
\newcite{xie2015modeling} use Unconditional Pure Dependence (UPD) to enhance QLM, leading to an improvement in information retrieval tasks. \newcite{zhang2018end} develop a model integrating word embeddings into density matrices to represent questions and answers, and the model is trained to compare questions and answers. \newcite{li2018quantum} propose a similar strategy in classification tasks. Among the existing studies, \newcite{sordoni2013modeling} and \newcite{xie2015modeling} focus on modeling the relations between questions and answers. They are interested in whether users and documents have quantum entanglements, and how to use quantum entanglements to improve the accuracy of information retrieval. 

It should be noted that all of the existing studies attempt to represent sentence-level information with density matrices.
If words are still represented by vectors, the advantage of sentence-level density matrices would be limited, and different aspects of word meaning could not be uncovered either. In addition, none of these works incorporate probability summation and vector summation between language units. Therefore, it is necessary to conduct a systematic study of using density matrices to represent language units from word level, and introduce probability summation into the computation between words.

\section{Conclusion and Future work}
\label{sec:conclusion}

In this paper, inspired by quantum probability, we represent words as density matrices, and successfully apply it to neural network models. This representation can effectively capture different aspects of word meaning while maintaining comparable reliability with the vector representation. In addition, we introduce two basic operations of quantum probability, i.e. probability summation and vector summation, into the computation between words in both vector form and density matrix form. This computation method could better sketch the morphological and semantic relations between words.
With a novel representation method and two operations, this work sheds some light on different mathematical forms of language representation.

Language is a combinatorial system, thus the representations of higher level language units e.g. sentences or texts are mostly built upon the computation of word meaning in a specific way. Meanwhile, different NLP tasks rely on different aspects of these representations. Therefore, in the future, we will investigate the density matrix representation of sentences or texts based on the word representations, and apply them to NLP downstream tasks in more complex neural networks.

\bibliographystyle{coling}
\bibliography{coling2020}

\begin{thebibliography}{}

\bibitem[\protect\citename{Devlin \bgroup et al.\egroup }2018]{devlin2018bert}
Jacob Devlin, Ming-Wei Chang, Kenton Lee, and Kristina Toutanova.
\newblock 2018.
\newblock Bert: Pre-training of deep bidirectional transformers for language
  understanding.
\newblock {\em arXiv preprint arXiv:1810.04805}.

\bibitem[\protect\citename{Dirac}1939]{dirac1939new}
Paul Adrien~Maurice Dirac.
\newblock 1939.
\newblock A new notation for quantum mechanics.
\newblock In {\em Mathematical Proceedings of the Cambridge Philosophical
  Society}, volume~35, pages 416--418. Cambridge University Press.

\bibitem[\protect\citename{Dirac}1942]{dirac1942bakerian}
Paul Adrien~Maurice Dirac.
\newblock 1942.
\newblock Bakerian lecture-the physical interpretation of quantum mechanics.
\newblock {\em Proceedings of the Royal Society of London. Series A.
  Mathematical and Physical Sciences}, 180(980):1--40.

\bibitem[\protect\citename{Fano}1957]{fano1957description}
Ugo Fano.
\newblock 1957.
\newblock Description of states in quantum mechanics by density matrix and
  operator techniques.
\newblock {\em Reviews of Modern Physics}, 29(1):74.

\bibitem[\protect\citename{Feynman}1987]{feynman1987negative}
Richard~P Feynman.
\newblock 1987.
\newblock Negative probability.
\newblock {\em Quantum implications: essays in honour of David Bohm}, pages
  235--248.

\bibitem[\protect\citename{Gladkova \bgroup et al.\egroup
  }2016]{gladkova2016analogy}
Anna Gladkova, Aleksandr Drozd, and Satoshi Matsuoka.
\newblock 2016.
\newblock Analogy-based detection of morphological and semantic relations with
  word embeddings: what works and what doesn’t.
\newblock In {\em Proceedings of the NAACL Student Research Workshop}, pages
  8--15.

\bibitem[\protect\citename{Levy and Goldberg}2014]{levy2014linguistic}
Omer Levy and Yoav Goldberg.
\newblock 2014.
\newblock Linguistic regularities in sparse and explicit word representations.
\newblock In {\em Proceedings of the eighteenth conference on computational
  natural language learning}, pages 171--180.

\bibitem[\protect\citename{Levy \bgroup et al.\egroup }2015]{levy2015improving}
Omer Levy, Yoav Goldberg, and Ido Dagan.
\newblock 2015.
\newblock Improving distributional similarity with lessons learned from word
  embeddings.
\newblock {\em Transactions of the Association for Computational Linguistics},
  3:211--225.

\bibitem[\protect\citename{Li \bgroup et al.\egroup }2018]{li2018quantum}
Qiuchi Li, Sagar Uprety, Benyou Wang, and Dawei Song.
\newblock 2018.
\newblock Quantum-inspired complex word embedding.
\newblock {\em arXiv preprint arXiv:1805.11351}.

\bibitem[\protect\citename{Mikolov \bgroup et al.\egroup
  }2013a]{mikolov2013efficient}
Tomas Mikolov, Kai Chen, Greg Corrado, and Jeffrey Dean.
\newblock 2013a.
\newblock Efficient estimation of word representations in vector space.
\newblock {\em arXiv preprint arXiv:1301.3781}.

\bibitem[\protect\citename{Mikolov \bgroup et al.\egroup
  }2013b]{mikolov2013distributed}
Tomas Mikolov, Ilya Sutskever, Kai Chen, Greg~S Corrado, and Jeff Dean.
\newblock 2013b.
\newblock Distributed representations of words and phrases and their
  compositionality.
\newblock In {\em Advances in neural information processing systems}, pages
  3111--3119.

\bibitem[\protect\citename{Pennington \bgroup et al.\egroup
  }2014]{pennington2014glove}
Jeffrey Pennington, Richard Socher, and Christopher Manning.
\newblock 2014.
\newblock Glove: Global vectors for word representation.
\newblock In {\em Proceedings of the 2014 conference on empirical methods in
  natural language processing (EMNLP)}, pages 1532--1543.

\bibitem[\protect\citename{Peters \bgroup et al.\egroup }2018]{peters2018deep}
Matthew~E Peters, Mark Neumann, Mohit Iyyer, Matt Gardner, Christopher Clark,
  Kenton Lee, and Luke Zettlemoyer.
\newblock 2018.
\newblock Deep contextualized word representations.
\newblock In {\em Proceedings of NAACL-HLT}, pages 2227--2237.

\bibitem[\protect\citename{Rogers \bgroup et al.\egroup }2018]{rogers2018s}
Anna Rogers, Shashwath~Hosur Ananthakrishna, and Anna Rumshisky.
\newblock 2018.
\newblock What’s in your embedding, and how it predicts task performance.
\newblock In {\em Proceedings of the 27th International Conference on
  Computational Linguistics}, pages 2690--2703.

\bibitem[\protect\citename{Sordoni \bgroup et al.\egroup
  }2013]{sordoni2013modeling}
Alessandro Sordoni, Jian-Yun Nie, and Yoshua Bengio.
\newblock 2013.
\newblock Modeling term dependencies with quantum language models for ir.
\newblock In {\em Proceedings of the 36th international ACM SIGIR conference on
  Research and development in information retrieval}, pages 653--662. ACM.

\bibitem[\protect\citename{Xie \bgroup et al.\egroup }2015]{xie2015modeling}
Mengjiao Xie, Yuexian Hou, Peng Zhang, Jingfei Li, Wenjie Li, and Dawei Song.
\newblock 2015.
\newblock Modeling quantum entanglements in quantum language models.
\newblock In {\em Twenty-Fourth International Joint Conference on Artificial
  Intelligence}.

\bibitem[\protect\citename{Zhang \bgroup et al.\egroup }2018]{zhang2018end}
Peng Zhang, Jiabin Niu, Zhan Su, Benyou Wang, Liqun Ma, and Dawei Song.
\newblock 2018.
\newblock End-to-end quantum-like language models with application to question
  answering.
\newblock In {\em Thirty-Second AAAI Conference on Artificial Intelligence}.

\end{thebibliography}

\end{document}